\documentclass{article}
\usepackage{amsmath}
\usepackage{amssymb}
\usepackage{pifont}
\usepackage{graphicx}
\usepackage{subcaption}
\usepackage{hyperref}
\usepackage{caption}
\usepackage{listings}
\usepackage{tablefootnote}
\usepackage{float}

\title{PathBench-MIL: A Comprehensive AutoML and Benchmarking Framework for Multiple Instance Learning in Histopathology}

\author{
    \textbf{Siemen Brussee}\textsuperscript{1} \thanks{Department of Pathology, Leiden University Medical Center. Contact: \href{mailto:s.brussee@lumc.nl}{s.brussee@lumc.nl}} \and
    Pieter A. Valkema\textsuperscript{1} \and
    Jurre A. J. Weijer\textsuperscript{1} \and
    Thom Doeleman\textsuperscript{1,2} \and
    Anne M.R. Schrader\textsuperscript{1} \and
    Jesper Kers\textsuperscript{1,3} 
}
\date{
    \textsuperscript{1}Department of Pathology, Leiden University Medical Center \\
    \textsuperscript{2}Department of Pathology, Utrecht University Medical Center
    \textsuperscript{3}Department of Pathology, Amsterdam University Medical Center \\
    19-12-2025
}
\begin{document}
\maketitle

\begin{abstract}
We introduce PathBench-MIL, an open-source AutoML and benchmarking framework for multiple instance learning (MIL) in histopathology. The system automates end-to-end MIL pipeline construction, including preprocessing, feature extraction, and MIL-aggregation, and provides reproducible benchmarking of dozens of MIL models and feature extractors. PathBench-MIL integrates visualization tooling, a unified configuration system, and modular extensibility, enabling rapid experimentation and standardization across datasets and tasks. PathBench-MIL is publicly available at \href{https://github.com/Sbrussee/PathBench-MIL}{https://github.com/Sbrussee/PathBench-MIL}.
\end{abstract}

\section{Introduction}
\subsection{Background}
Computational pathology has grown rapidly due to the increasing availability of digitized whole-slide images (WSI) and advances in computational power. 

The gigapixel scale of WSIs presents a significant obstacle, as these massive images do not typically fit into the memory of graphical processing units (GPUs), which are necessary for the efficient training of deep learning models. Simultaneously, the high cost of obtaining labeled annotations, which often necessitates expert pathologists, limits the availability of fine-grained, pixel- or region-level labels. As a result, most labels are available only at the slide level, leading to the need for \textit{weakly supervised} machine learning approaches that can effectively link local image patterns with global slide-level labels.

One widely adopted weakly supervised approach to address these challenges is \textit{multiple instance learning} (MIL), which has emerged as the predominant paradigm for AI modeling in computational pathology \cite{gadermayr2024multiple, doeleman2024deep} (see Figure \ref{fig:mil_pipeline}). MIL methods subdivide WSIs into smaller, manageable patches (or instances), enabling the model to process image data in a computationally feasible way. In MIL, a bag of instances (patches) represents each slide, and only the slide-level label is required. The model learns to associate the most informative patches with the global label, bypassing the need for costly, fine-grained annotations. A more formal description of the MIL formulation is provided in Appendix~\ref{appendix:mil}.

\subsection{Motivation: The MIL pipeline search problem}
\begin{figure}[H]
    \centering
    \includegraphics[width=1.0\linewidth]{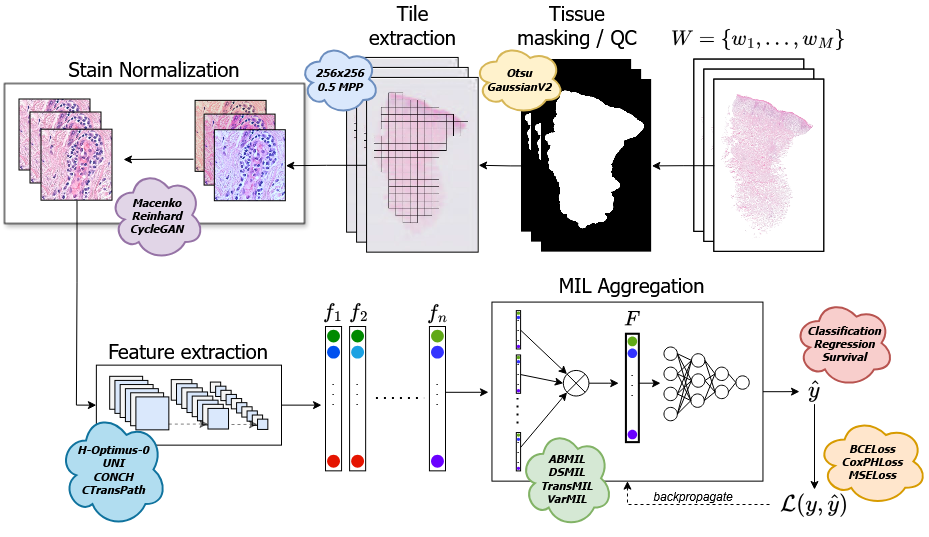}
    \caption{Modular MIL pipeline illustrating the combinatorial search space of histopathology MIL frameworks. Variations in preprocessing, tiling, stain normalization, feature extractors (including foundation models), and aggregation strategies yield a large set of candidate pipelines that PathBench-MIL benchmarks and optimizes end-to-end.}
    \label{fig:mil_pipeline}
\end{figure}

Strong MIL performance depends on coherent design choices across the full pipeline, including tiling, tissue segmentation, artifact filtering, stain handling, feature extraction (increasingly via foundation models), aggregation architecture, and training objectives. Because each stage admits multiple valid alternatives, finding optimal MIL pipeline configurations in computational pathology has become a combinatorial \emph{search space} problem (Figure \ref{fig:mil_pipeline}).

This expanding design space is difficult to explore manually: evaluating pipeline configurations across multiple datasets is computationally expensive, and optimal configurations are often dataset- and task-dependent. Crucially, pipeline components interact, so gains from improving a single stage may not translate to improved slide-level performance without compatible choices elsewhere.

Benchmarking studies provide useful insight into the effect of MIL pipeline configurations, but most comparisons focus on individual stages (e.g., feature extraction \cite{kang2023benchmarking,campanella2024clinical,neidlinger2024benchmarking}, aggregation \cite{laleh2022benchmarking,chen2024benchmarking} or stain normalization/augmentation \cite{tellez2019quantifying}) rather than end-to-end pipelines, limiting insight into how preprocessing, feature extraction, and MIL architecture jointly drive performance.

Automated Machine Learning (AutoML) \cite{waring2020automated} offers a principled way to address the design space problem by systematically searching over discrete pipeline choices and continuous hyperparameters in a reproducible, configuration-driven manner. Joint optimization across stages is therefore well-suited to identifying robust, high-performing MIL pipelines in computational pathology.

\subsection{Introducing PathBench-MIL}
\begin{figure}[H]
    \centering
    \includegraphics[width=1.0\linewidth]{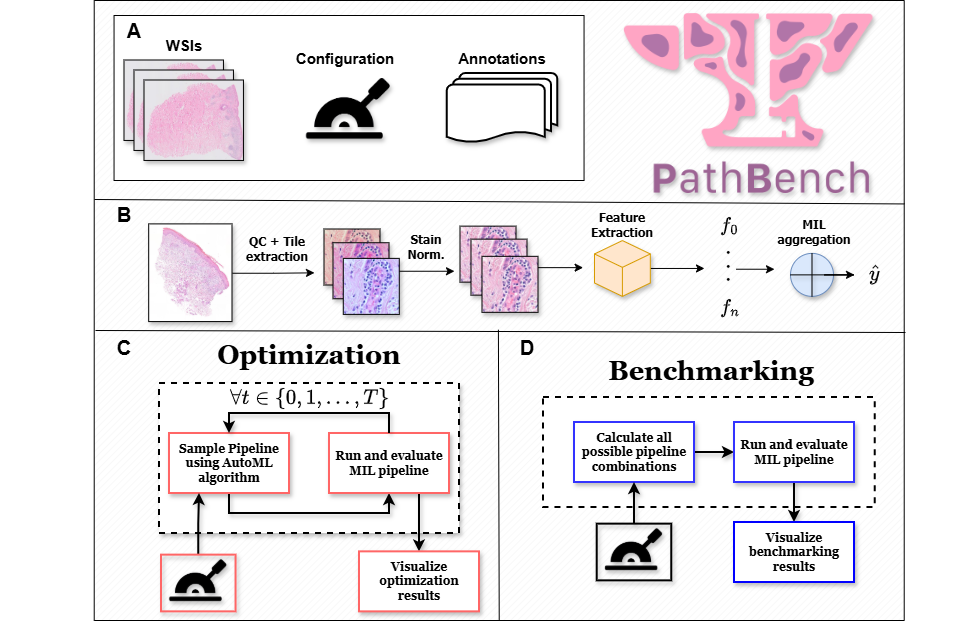}
        \caption{
    Overview of the PathBench-MIL framework. 
    \textbf{(A)} Required inputs: whole-slide images (WSIs), a configuration file specifying 
    pipeline components, and slide-level annotations. 
    \textbf{(B)} The end-to-end MIL pipeline, including quality control (QC), tile extraction, 
    stain normalization, tile-level feature extraction, and slide-level aggregation to produce 
    final predictions. 
    \textbf{(C)} Optimization mode: an AutoML engine samples full pipeline configurations, 
    evaluates each configuration, and stores results for interactive visualization. 
    \textbf{(D)} Benchmarking mode: all user-specified pipeline combinations are enumerated 
    and evaluated, enabling systematic comparison across MIL design choices.}
    \label{fig:pathbench_ga}
\end{figure}

To address the need for end-to-end evaluation and scalable pipeline optimization, we introduce \textbf{PathBench-MIL}, a flexible benchmarking and AutoML framework designed specifically for multiple instance learning in pathology (Figure \ref{fig:pathbench_ga}). PathBench-MIL is built on top of SlideFlow \cite{dolezal2024slideflow}, leveraging its efficient WSI data handling, multiprocessing support, and unified data-source abstraction. It extends this foundation with full pipeline benchmarking, automated pipeline search, and dedicated tooling for survival tasks and interactive visualization.

PathBench-MIL makes three main contributions:
\begin{itemize}
    \item An end-to-end benchmarking framework that evaluates complete MIL pipelines rather than isolated components.
    \item An AutoML optimization engine that searches over both numeric hyperparameters and discrete pipeline design choices specific to pathology MIL.
    \item Integrated analysis and visualization tools for interactive exploration of large benchmarking and AutoML result spaces.
\end{itemize}

\section{PathBench-MIL Framework}
PathBench-MIL is an open-source software framework for end-to-end benchmarking and automated optimization of multiple instance learning (MIL) pipelines in computational histopathology. It provides a unified configuration interface for the core pipeline stages—preprocessing (including tiling and optional stain handling), tile-level feature extraction, and slide-level aggregation—and supports classification, regression, and both continuous and discrete survival prediction tasks. In addition, PathBench-MIL includes integrated visualization utilities for comparative analysis and model inspection across large experimental result spaces.

\subsection{Operational Modes}
PathBench-MIL provides two primary operational modes, \textit{Benchmarking} and \textit{Optimization}, corresponding to different stages of model development and evaluation.

\subsubsection{Benchmarking Mode}
Benchmarking mode enables systematic comparison of full MIL pipelines across user-specified datasets and parameter configurations. Users may define multiple datasets for training, validation, and external testing to assess model generalization. For each configuration, PathBench-MIL evaluates the complete pipeline and reports task-appropriate performance metrics.

Key configurable components include:
\begin{itemize}
    \item \textbf{Tiling Parameters:} Patch size and magnification.
    \item \textbf{Stain Normalization:} Methods such as Macenko and Reinhard.
    \item \textbf{Feature Extractors:} For example, ResNet50, UNI~\cite{chen2024towards}, or PLIP~\cite{huang2023visual}.
    \item \textbf{Aggregation Models:} MIL architectures including CLAM~\cite{lu2021data}, DSMIL~\cite{li2021dual}, and AB-MIL~\cite{ilse2018attention}.
    \item \textbf{Loss Functions and Optimizers:} User-defined choices for training objectives and optimization algorithms.
\end{itemize}

Benchmarking mode evaluates all specified configurations and records results in a structured format compatible with downstream visualization.

\subsubsection{Optimization (AutoML) Mode}
Optimization mode provides an end-to-end AutoML engine for MIL pipeline design, implemented on top of the Optuna framework~\cite{akiba2019optuna}. The engine performs a budget-constrained, guided search over a structured pipeline space, replacing exhaustive benchmarking with adaptive sampling and early stopping. Users may select among multiple samplers (e.g., TPE~\cite{bergstra2011algorithms}, CMA-ES, Random Search) and pruning strategies (e.g., Hyperband~\cite{li2018hyperband}, Median Pruning), which together balance exploration of the search space with exploitation of configurations that demonstrate promising intermediate performance, ultimately saving compute time compared to benchmarking all possible configurations.

Concretely, when combining TPE sampling with Hyperband pruning, the expected wall-clock speedup relative to exhaustive evaluation over $N$ pipeline configurations can be approximated as
\[
\text{Speedup} \;\approx\; \frac{N}{T} \cdot \frac{1}{f},
\]
where $T$ denotes the number of optimization trials and $f \in [0.05, 0.3]$ is the average fraction of full training cost incurred per trial due to early termination. Empirical studies show that Hyperband alone can achieve over an order-of-magnitude ($\sim$10$\times$) reduction in time-to-good-solution, while TPE further improves sample efficiency in structured and conditional search spaces, yielding practical speedups of one to two orders of magnitude compared to full benchmarking~\cite{bergstra2011algorithms,li2018hyperband,falkner2018bohb}.

To support large-scale optimization on HPC systems, trial states and artifacts are persisted automatically, enabling seamless recovery from preemption and robust execution under batch scheduling. Additional details on the trial lifecycle and pruning behavior are provided in Appendix~\ref{appendix:optimization}.

PathBench-MIL models the MIL pipeline as a structured configuration space spanning preprocessing, feature extraction, aggregation, and training hyperparameters. Let $\mathcal{S}$ denote the global search space, decomposed as
\begin{equation}
    \mathcal{S} =
    \mathcal{T}_{tile} \times
    \mathcal{P}_{norm} \times
    \mathcal{M}_{feat} \times
    \mathcal{A}_{agg} \times
    \mathcal{H}_{opt}.
\end{equation}

Each configuration $\mathbf{c} \in \mathcal{S}$ defines a complete MIL pipeline. The AutoML objective is to identify
\begin{equation}
    \mathbf{c}^* =
    \arg\min_{\mathbf{c} \in \mathcal{S}}
    \mathcal{L}_{val}\!\left(
        \text{Train}(\mathbf{c}, \mathcal{D}_{train}),
        \mathcal{D}_{val}
    \right),
\end{equation}
where $\mathcal{L}_{val}$ denotes the validation objective (e.g., AUC).

To further reduce redundant computation in large optimization studies, PathBench-MIL employs a caching mechanism: if a preprocessing subconfiguration $\mathbf{c}_{pre} \subset \mathbf{c}$ has been executed previously, its associated artifact $\Phi(\mathbf{c}_{pre})$ (e.g., tiles or feature embeddings) is reused across trials. This substantially reduces I/O overhead and accelerates iterative AutoML workflows. 

\subsection{Design and Implementation}
\subsubsection{User Workflow and Extensibility}
PathBench-MIL is designed with reproducibility and usability as core principles. All stages of the pipeline—including tiling parameters, stain normalization, feature extraction, MIL aggregation, and optimization settings—are specified in a single, unified YAML configuration file. This centralized configuration ensures transparent experiment definition and facilitates straightforward replication across users and compute environments.

The framework follows a fully modular architecture. Core components such as feature extractors, MIL aggregators, loss functions, data augmentation strategies, and evaluation metrics are implemented via standardized interfaces, allowing new methods to be integrated with minimal effort. This design enables systematic benchmarking of emerging models against established baselines and supports adaptation to evolving computational pathology workflows.

Using PathBench-MIL requires only three inputs: (i) a valid YAML configuration file defining the experimental setup, (ii) a CSV file containing slide-level annotations (including at minimum slide ID, patient ID, and prediction label), and (iii) the corresponding whole-slide images. Together, these inputs are sufficient to execute end-to-end experiments in a reproducible and extensible manner. For more detailed usage instructions, we refer to the PathBench-MIL \href{https://github.com/Sbrussee/PathBench-MIL}{GitHub repository}.

\subsubsection{Visualization Application}
To support interpretation of benchmarking and AutoML results, PathBench-MIL includes an interactive visualization application built with the Dash framework. Users can explore performance metrics across all evaluated pipeline parameters, create custom plots, filter or group results, and export figures for downstream analysis. This interface provides practical insight into parameter interactions and model behavior, facilitating informed pipeline selection.

\subsection{Related Work \& Comparison}
Several benchmarking frameworks have been introduced for computational pathology, with a predominant focus on evaluating foundation models rather than full pipeline configurations. Patho-Bench, introduced by Zhang et al.~\cite{zhang2025standardizing}, provides large-scale evaluations of pathology foundation models across numerous tasks using standardized dataset splits. Similarly, EVA~\cite{kaiko.ai2024eva} offers a comprehensive evaluation environment tailored to oncology foundation models, supporting tasks such as classification, segmentation, and embedding-based analyses.

These frameworks, however, do not evaluate full MIL pipelines end-to-end. They primarily assess foundation model representations rather than combinations of preprocessing, tiling, feature extraction, aggregation, and optimization choices. PathBench-MIL fills this gap by providing a software system for holistic MIL pipeline evaluation and automated pipeline search.

PathBench-MIL differs in scope by providing an extensible software framework for holistic MIL pipeline evaluation. It supports both continuous and discrete deep survival analysis, exposes AutoML capabilities for joint pipeline configuration search, and includes interactive visualization tools for inspecting benchmarking or optimization results. These features enable systematic experimentation across all stages of MIL workflows, complementing the foundation-model–focused benchmarking provided by Patho-Bench and EVA.

Table~\ref{tab:benchmark_comparison} summarizes key differences between PathBench-MIL and existing frameworks.

\begin{table}[H]
    \centering
    \caption{Comparison of PathBench-MIL, Patho-Bench, and EVA.}
    \label{tab:benchmark_comparison}
    \begin{tabular}{|l|c|c|c|}
        \hline
        \textbf{Feature} & \textbf{PathBench-MIL} & \textbf{Patho-Bench} & \textbf{EVA} \\
        \hline
        Covers entire MIL pipeline & \ding{51} & \ding{55} & \ding{55} \\ 
        Patch foundation models & \ding{51} & \ding{51} & \ding{51} \\
        Slide-level foundation models & \ding{51} & \ding{51} & \ding{55} \\
        Classification & \ding{51} & \ding{51} & \ding{51} \\
        Regression & \ding{51} & \ding{55} & \ding{55} \\
        Deep continuous survival & \ding{51} & \ding{55} & \ding{55} \\
        Deep discrete survival & \ding{51} & \ding{51} & \ding{55} \\
        AutoML capabilities & \ding{51} & \ding{55} & \ding{55} \\
        Interactive visualization & \ding{51} & \ding{55} & \ding{55} \\
        CPU multiprocessing & \ding{51} & \ding{51} & \ding{51} \\
        GPU parallelization & \ding{51} & \ding{51} & \ding{55} \\
        \hline
        Built-in datasets \& tasks & \ding{55} & \ding{51} & \ding{51} \\
        Semantic segmentation & \ding{55} & \ding{55} & \ding{51} \\
        Patch-level prediction tasks & \ding{55} & \ding{55} & \ding{51} \\
        Retrieval & \ding{55} & \ding{51} & \ding{55} \\
        \hline
    \end{tabular}
\end{table}

\section{Discussion}
\par{We presented PathBench-MIL, an end-to-end AutoML and benchmarking framework for multiple instance learning in computational pathology. We described the framework's benchmarking and AutoML capabilities, its interactive visualization capabilities and its extendibility.}

\subsection{Broader Impact}
\par{
Beyond methodological contributions, PathBench-MIL may also reduce the engineering burden of MIL pipeline development for clinical and translational researchers. By providing standardized end-to-end benchmarking and AutoML-driven pipeline search, the framework can help users with limited machine learning infrastructure or experience to more systematically evaluate modern MIL components and foundation-model-based feature extractors, while focusing on dataset curation and clinical interpretation.

}
\subsection{Limitations}
\par{One of the primary challenges when conducting large-scale benchmarking or optimization tasks is the computational cost, which scales combinatorially with the cardinality of the search space. PathBench-MIL does not eliminate the computational cost inherent to large-scale benchmarking. While caching and early pruning reduce redundant work, users may still need substantial compute for exhaustive studies.}

\subsection{Future Directions}
\par{We envision several key developments to further enhance the utility and scalability of PathBench-MIL. First, regarding infrastructure, we aim to facilitate reproducibility by integrating direct download capabilities for public datasets. Additionally, to accommodate the growing computational demands of large-scale studies, we plan to implement distributed benchmarking, allowing users to parallelize experiments across multiple GPUs \cite{dally2021evolution}.}

\par{Second, we intend to expand the framework's architectural scope beyond standard MIL. This includes support for \textit{multimodal modeling}, enabling the integration of clinical data or other imaging modalities to improve predictive performance. We also propose the inclusion of \textit{ensemble modeling} to combine diverse feature extractors and aggregators. Furthermore, we aim to introduce a dedicated benchmarking module for pathology-focused \textit{Graph Neural Networks (GNNs)} \cite{brussee2025graph}. By extending PathBench-MIL’s AutoML capabilities to graph-based approaches, the framework could automatically optimize upstream design choices—such as nuclei segmentation and graph construction algorithms—allowing researchers to identify the optimal topological representations for specific downstream tasks.}

\par{Finally, we plan to extend the framework's versatility by supporting a broader range of computational pathology tasks. Future updates will target capabilities for \textit{image retrieval} based on feature similarity \cite{chen2022fast}, \textit{morphological prototyping} via clustering, and dense prediction tasks such as nuclei and semantic segmentation. These additions will position PathBench-MIL as a holistic platform for developing and evaluating diverse pathology AI pipelines.}

\section{Conclusion}
\par{PathBench-MIL provides an extensible software framework for benchmarking and automated optimization of MIL pipelines in computational pathology. By integrating AutoML-driven search, end-to-end pipeline configuration, and interactive visualization, the framework reduces the time, expertise, and engineering effort required to build robust MIL models. We anticipate that PathBench-MIL will serve as a foundation for future methodological comparisons and standardized evaluation practices in pathology AI research.}

\section*{CRediT Author Statement}
\textbf{Conceptualization}: Siemen Brussee\\
\textbf{Methodology}: Siemen Brussee\\
\textbf{Software}: Siemen Brussee, Pieter A. Valkema, Jurre Weijer\\
\textbf{Validation}: Pieter A. Valkema, Jurre Weijer, Thom Doeleman\\
\textbf{Formal Analysis}: Siemen Brussee\\
\textbf{Investigation}: Siemen Brussee\\
\textbf{Resources}: Pieter A. Valkema, Thom Doeleman, Jesper Kers, Anne M.R. Schrader\\
\textbf{Data Curation}: Pieter A. Valkema, Thom Doeleman, Anne M.R. Schrader \\
\textbf{Writing - Original Draft}: Siemen Brussee\\
\textbf{Writing - Review \& Editing}: Pieter A. Valkema, Thom Doeleman, Jurre Weijer, Jesper Kers, Anne M.R. Schrader\\
\textbf{Visualization}: Siemen Brussee\\
\textbf{Supervision}: Jesper Kers, Anne M.R. Schrader\\
\textbf{Project Administration}: Jesper Kers, Anne M.R. Schrader\\
\textbf{Funding Acquisition}: Anne M.R. Schrader\\

\section*{Funding}
This work was supported by the Hanarth Foundation for AI in Oncology, The Netherlands.
\\ \\
\textbf{Keywords}: Multiple Instance Learning, Computational Pathology, Foundation Models, AutoML, Automated Machine Learning, Benchmarking, Hyperparameter Optimization, Histopathology
\bibliographystyle{unsrt}
\bibliography{bib}

\appendix
\section{MIL Background and Formulation}
\label{appendix:mil}

Multiple Instance Learning (MIL) provides a natural framework for whole-slide image (WSI) analysis, 
where each slide is represented as a bag of tiles rather than a single fixed-size image. Let 
$\mathcal{X} = \{ x_1, x_2, \dots, x_n \}$ denote the set of tile embeddings for a slide, and 
$y$ the corresponding slide-level label. MIL models typically follow a two-stage structure comprising 
(1) tile-level feature extraction and (2) slide-level aggregation.

Formally, a MIL model can be written as:
\begin{equation}
    \hat{y} = f_{\text{agg}}\!\left( g_{\text{feat}}(\mathcal{X}) \right),
\end{equation}
where $g_{\text{feat}}$ extracts feature embeddings for each tile and $f_{\text{agg}}$ aggregates 
these embeddings into a slide-level prediction. Aggregation functions include attention-based pooling, 
transformer-style models, and clustering-based mechanisms.

Given training data $\{(\mathcal{X}^{(i)}, y^{(i)})\}_{i=1}^N$, MIL aims to learn parameters 
$\theta$ minimizing a task-specific loss:
\begin{equation}
    \theta^* = \arg\min_{\theta} 
    \sum_{i=1}^N \mathcal{L}\!\left( 
        f_{\text{agg}}(g_{\text{feat}}(\mathcal{X}^{(i)};\,\theta)),\,
        y^{(i)}
    \right).
\end{equation}

This formulation applies to classification, regression, and continuous or discrete survival 
prediction. In PathBench-MIL, the full MIL pipeline—including tiling, stain normalization, feature 
extraction, aggregation, and optimization—can be expressed as a structured configuration, enabling 
systematic end-to-end evaluation or automated search over MIL architectures and hyperparameters.

\section{Optimization Workflow Details}
\label{appendix:optimization}

PathBench-MIL implements a robust optimization workflow built on Optuna to support large-scale 
AutoML studies in MIL. Each optimization trial proceeds through a standardized lifecycle:

\subsection*{Trial Lifecycle}
\begin{enumerate}
    \item \textbf{Sampling:} A sampler (e.g., TPE, CMA-ES, Random Search) proposes a configuration 
    comprising preprocessing steps, feature extractors, aggregation architectures, and training 
    hyperparameters.
    
    \item \textbf{Caching:} Before execution, PathBench-MIL checks whether artifacts corresponding to 
    any preprocessing subconfiguration exist (e.g., tiles or feature bags). If so, they are reused, 
    avoiding redundant computation and minimizing I/O overhead.
    
    \item \textbf{Evaluation:} The sampled pipeline is trained and validated, optionally using 
    cross-validation for more robust performance estimation.

    \item \textbf{Pruning:} A pruning strategy such as Hyperband monitors intermediate validation 
    metrics and terminates underperforming trials early. A warm-up phase ensures that a set of 
    initial trials runs to completion to establish a baseline before pruning is activated.
    
    \item \textbf{Persistence and Fault Tolerance:} Trial states are serialized after each iteration, 
    enabling studies to resume seamlessly after job preemption or wall-time limits on shared HPC 
    clusters.
\end{enumerate}

\subsection*{Pruning Strategy}
Hyperband organizes trials into rungs based on allocated training budgets. At predefined checkpoints, underperforming configurations are discarded while more promising ones continue to subsequent rungs. This structure enables significant acceleration of large design-space searches while maintaining reliable performance estimation.

\begin{figure}[H]
    \centering
    \begin{subfigure}[c]{0.35\textwidth}
        \centering
        \includegraphics[width=\linewidth]{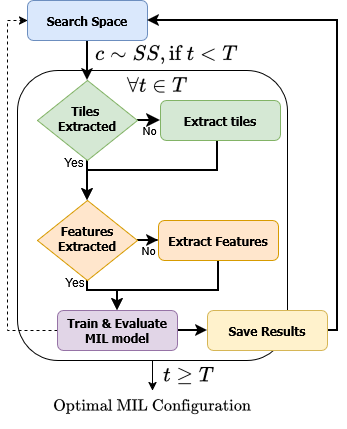}
        \caption{Optimization Workflow}
        \label{fig:automl_workflow}
    \end{subfigure}
    \hfill 
    \begin{subfigure}[c]{0.63\textwidth}
        \centering
        \includegraphics[width=\linewidth]{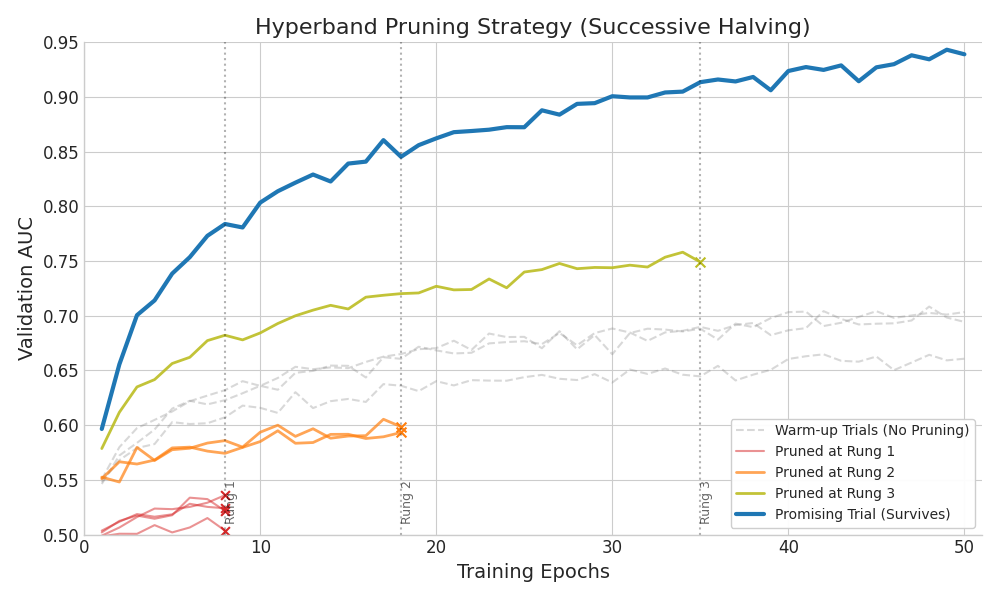}
        \caption{Hyperband Pruning Strategy}
        \label{fig:hyperband_plot}
    \end{subfigure}
\caption{\textbf{Efficiency mechanisms in PathBench-MIL.} 
    \textbf{(a)} The AutoML optimization workflow. For each trial $t$ within the total budget $T$ ($t < T$), a pipeline configuration $c$ is sampled from the search space $SS$ ($c \sim SS$). The framework utilizes conditional logic (diamond nodes) to check for existing data artifacts; if tiles or features for configuration $c$ already exist, extraction is skipped. The model is then trained and evaluated. If the trial is not pruned (dashed line), results are saved. This loop repeats until the budget is exhausted ($t \ge T$) to identify the optimal configuration.
    \textbf{(b)} The Hyperband pruning strategy visualizes the successive halving of trials. After a warm-up phase (gray dashed lines), underperforming trials (red/orange) are terminated at specific checkpoints ("rungs"), reserving resources for the most promising candidates (blue).}
    \label{fig:automl_combined}
\end{figure}

Figure~\ref{fig:automl_combined} illustrates the overall optimization workflow, including trial sampling, caching, and the hyperband pruning strategy.

\end{document}